%% file: main.tex
\documentclass{article}
\usepackage{spconf,amsmath,graphicx}
\usepackage{multirow}
\usepackage{xcolor}
\usepackage{xspace} 
\usepackage{soul}

\newcommand{\algname}{STHL\xspace}

\usepackage{caption}
\DeclareMathOperator*{\concat}{%
    \mathchoice%
        {\Big\Vert}%
        {\big\Vert}%
        {\Vert}%
        {\Vert}%
}

\title{Exploiting Spatial-temporal Data for Sleep Stage Classification via Hypergraph Learning}

\name{\normalsize{Yuze Liu}\textsuperscript{\rm 1}$^{\star}$\thanks{$^{\star}$Co-first Authorship.}\quad\normalsize{Ziming Zhao}\textsuperscript{\rm 1}$^{\star}$ \quad\normalsize{Tiehua Zhang}\thanks{$^{\dagger}$Corresponding Author.}\textsuperscript{\rm 1}$^{\dagger}$ \quad\normalsize{Kang Wang\textsuperscript{\rm 2}}\quad\normalsize{Xin Chen}\textsuperscript{\rm 1}\quad\normalsize{Xiaowei Huang}\textsuperscript{\rm 1}\quad\normalsize{Jun Yin}\textsuperscript{\rm 1}\quad\normalsize{Zhishu Shen}\textsuperscript{\rm 2}$^{\dagger}$}
\address{\textsuperscript{\rm 1}\normalsize{Ant Group, Hangzhou, China}\\
\textsuperscript{\rm 2}\normalsize{School of Computer Science and Artificial Intelligence, Wuhan University of Technology, Wuhan, China}}

\begin{document}
%
\maketitle
\begin{abstract}
Sleep stage classification is crucial for detecting patients' health conditions. Existing models, which mainly use Convolutional Neural Networks (CNN) for modelling Euclidean data and Graph Convolution Networks (GNN) for modelling non-Euclidean data, are unable to consider the heterogeneity and interactivity of multimodal data as well as the spatial-temporal correlation simultaneously,  which hinders a further improvement of classification performance. In this paper, we propose a dynamic learning framework \algname, which introduces hypergraph to encode spatial-temporal data for sleep stage classification. Hypergraphs can construct multi-modal/multi-type data instead of using simple pairwise between two subjects. \algname creates spatial and temporal hyperedges separately to build node correlations, then it conducts type-specific hypergraph learning process to encode the attributes into the embedding space. Extensive experiments show that our proposed \algname outperforms the state-of-the-art models in sleep stage classification tasks.
\end{abstract}

\input{1_introduction}

\input{2_related_work}

\input{3_methodology}

\input{4_experiment}

\input{5_conclusion}

\clearpage
\bibliographystyle{IEEEbib}
\bibliography{refs}

\end{document}

%% file: 1_introduction.tex
\section{Introduction}
\label{sec:intro}

Scientific sleep stage classification is of significance to the diagnosis of sleep disorders to enhance the sleep quality. As a key reference, polysomnography (PSG) is utilised to classify the sleep stages. 
PSG records the physical signals of different organs of the human body. These signals include electroencephalography (EEG), electrooculography (EOG), electromyography (EMG), and electrocardiography (ECG). The interactions between organs and the influences of adjacent timestamps render the observations non-independent. Therefore, the data recorded by PSG is considered \textit{spatial-temporal}~\cite{101609, Yu_2018}, which is associated with time series and spatial geographic location. 
Meanwhile, substantial differences exist in the spectrograms recorded from multiple organs in PSG, and thus it is also crucial to utilise the multimodal signals originated from different signal source types, e.g., EOG signals and ECG signals. Besides the spatial-temporal correlations of the data, two properties also need to be considered when handling these multimodal signals for sleep stage classification:
the \textit{interactivity} that indicates the continuous interaction of human organs during sleep periods \cite{9287428}, and the \textit{heterogeneity} that represents the difference in spectrograms exhibited when detecting different signals ~\cite{8512585}.

Recently, deep learning based methods \cite{CR}, such as Convolutional Neural Networks (CNN) and Recurrent Neural Network (RNN), have been widely studied to solve the sleep stage classification problem due to their powerful representation learning capabilities. However, the process of CNN/RNN requires the image data to be represented as a regular grid-like (Euclidean) data structure, i.e., a grid of pixels. Apart from that, it is pointed out that prior models fail to capture the implicit correlations in the data simultaneously~\cite{SleepHGNN}. To alleviate this limitation, Graph Neural Network (GNN), which is a neural network designed for handling graph data, has garnered significant attention in addressing the generation of data from non-Euclidean domains~\cite{Ahmedt_Aristizabal_2021}. Moreover, hypergraph-based methods start to emerge to process multimodal data efficiently~\cite{9795251}. Compared with the traditional graph learning-based method where each edge connects only two nodes, hypergraphs introduce multiple \textit{hyperedges} that can connect more than two nodes
simultaneously. Therefore, hypergraph is expected to extract high-order correlations from multimodal signals, leading to a superior classification performance.

In this paper, we propose a dynamic learning framework, namely \algname, to achieve accurate sleep stage classification. To the best of our knowledge, this is the first time that hypergraph is utilised to analyse the multimodal signals for sleep stage classification. Our main contributions are summarised as follows: (1) We propose a hypergraph-based framework for sleep stage classification. It includes dynamic hyperedge construction and embedding update and multi-head attentive node embedding update. (2) We design a dynamic learning process to generate spatial and temporal hyperedges separately, forming the hypergraph with better modelling capability. The node embedding is then updated through multi-head attention mechanism to encode the interactivity and heterogeneity into the embedding spaces. (3) We conduct extensive experiments using the real-world dataset, demonstrating that our proposal can outperform comparative methods in sleep stage classification.

%% file: 2_related_work.tex
\section{Related work}
\subsection{Traditional Learning Methods}
 Traditional machine learning-based sleep stage classification, such as Support Vector Machine (SVM) \cite{HASSAN2016107}, Random Forest (RF) \cite{RF}, and Hidden Markov Model (HMM) \cite{HMM} not only rely heavily on the quality of feature extraction but also fail to extract important spatial-temporal relationships. 
On the other hand, while CNN \cite{CNN} and RNN \cite{CR}, which are proven effective when extracting the features to improve the classification, these methods rely on grid data input, and thus lack the ability to capture the connections between different brain regions and organs. As a result, incorporating graphs as representative data structures for connections in non-Euclidean spaces is then investigated for accurate modelling. 
\subsection{Graph-based Learning Methods}
Jia \textit{et al.}~\cite{GraphSleepNet} proposed GraphSleepNet that introduces a spatial-temporal graph convolutional network to learn the intrinsic connection between different EEG signals adaptively. Developed from GraphSleepNet, two brain views are constructed based on the functional connectivity and physical distance proximity of brain regions, by which a multi-view spatial-temporal map convolutional network (MSTGCN)~\cite{MSTGCN} is constituted. Moreover, SleepHGNN~\cite{SleepHGNN} combines the characteristics of heterogeneity and interactivity, while considering the interactions between human organs. However, dynamic spatial-temporal correlation of the data is not included in SleepHGNN.

In light of using attention mechanisms to capture the information between sleep stages~\cite{MGT,AGS}. As aforementioned, hypergraphs combine multiple hyperedges with different types for representing higher-order interrelations, and thus it is effective in solving the association problem of multimodal data. In this paper, we introduce the hypergraph-based framework for sleep stage classification, and the hypergraph is constructed dynamically so that implicit spatial-temporal correlations of the data can be encoded.
\label{sec:related}

%% file: 3_methodology.tex
\begin{figure*}[tb!]
    \centering
    \includegraphics[width=1\linewidth,height = 0.24\textwidth]{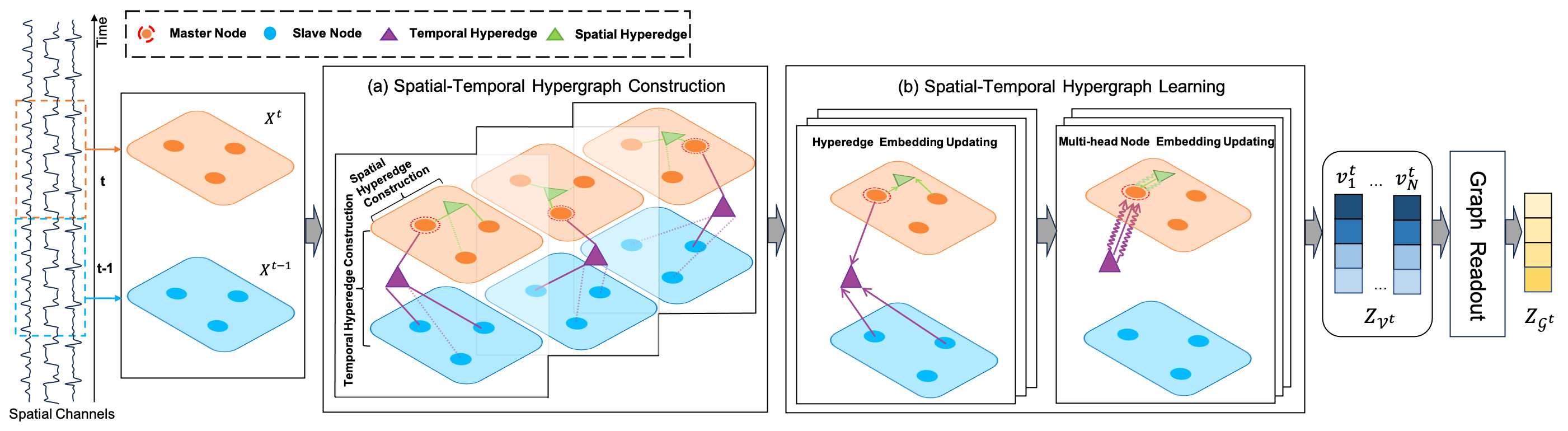}
  \caption{An overview of \algname}
  \label{fig:all} 
\end{figure*} 
\section{methodology}
\label{sec:meth}
Let $\mathcal{X}\!=\!\left\{\emph{X}^{1},...,\emph{X}^{T}\right\}\in\mathcal{R}^{T\times N\times d}$ denotes a spatial-temporal data containing $T$ timestamps, where $\emph{X}^{t}\!=\!\left\{\emph{x}^{t}_{1},...,\emph{x}^{t}_{N}\right\}\in \mathcal{R}^{N\times d}$ represents the signal of $N$ channels at a timestamp $t$, and the size of each channel feature is $d$ (As shown in the leftmost part of Fig.~\ref{fig:all}). We define a hypergraph as $\mathcal{G} = \left\{\mathcal{V},\mathcal{E}\right\}$, where $\mathcal{V}$ is the set of node set and $\emph{E}$ is the set of hyperedges. Each hyperedge $e\in\mathcal{E}$ contains two or more nodes. Since there is no node concept in the spatial-temporal data $\mathcal{X}$, we thus treat each spatial channel as a node at any timestamp $t$. The node set at timestamp $t$ is defined as $\emph{V}^{t} = \left\{v^{t}_{1},...,v^{t}_{N}\right\}$, and the number of nodes in each timestamp is $N$. We use $\emph{Y} = \left\{\emph{y}^{1},...,\emph{y}^{T}\right\}$ to denote the label of each timestamp. 

\subsection{Spatial-Temporal Hypergraph Construction}
The spatial-temporal hypergraph at timestamp $t$ is defined as $\mathcal{G}^{t} = \left\{
\mathcal{V}^{t},\mathcal{E}^{t}\right\}$. To model the temporal dependencies among nodes along the time dimension, we utilise the nodes feature of one timestamp previous to capture the historical information \cite{zhang2022fedrel}. Thus, the node set $\mathcal{V}^{t}$ contains nodes from two consecutive timestamps at $t$, which is defined as $\mathcal{V}^{t} = \left\{\emph{V}^{t},\emph{V}^{t-1}\right\}$. The node feature matrices for these timestamps are $\emph{X}^{t}$ and $\emph{X}^{t-1}$ respectively.

A hyperedege encloses a specific set of nodes with common attributes or implicit data relations, which represents local group information among the nodes. To facilitate hyperedge generation, we introduce two kinds of nodes: master node and slave node. A master node $\dot{v}\in\mathcal{V}$ acts as an anchor when generating the hyperedge $e\left(\dot{v}\right)\in\mathcal{E}$, which is combined with a set of slave nodes $S\left(\dot{v}\right) = \left\{\hat{v}\right\}$ to collectively constitute the hyperedge. We propose a dynamic learning process that generates two types of hyperedges: spatial hyperedges that capture heterogeneity, encoding relations among different channels in one timestamp; temporal hyperedges that scrutinize interactivity,  modelling the continuous interaction of channels in consecutive timestamps.

For each master node $\dot{v}^{t}_{i}\in\emph{V}^{t}$ in spatial-temporal hypergraph $\mathcal{G}^{t}$ at timestamp $t$, 
the spatial hyperedge $e_{spa}\left(\dot{v}_{i}^{t}\right)\in\mathcal{E}^{t}$ can be generated based on the reconstruction of the master node $\dot{v}^{t}_{i}$ and the spatial candidate slave node set $\Tilde{S}^{spa}\left(\dot{v}^{t}_{i}\right) = \left\{v|v\in\emph{V}^{t},v\neq\dot{v}^{t}_{i}\right\}$, which is denoted as:
\begin{small}
    \begin{equation}
    c_{spa}\left(\dot{v}^{t}_{i}\right)= \|\emph{X}^{t}\left(\dot{v}^{t}_{i}\right)\cdot\theta_{spa}-\emph{p}^{spa}_{\dot{v}^{t}_{i}}\cdot\emph{X}^{t}\left(\Tilde{S}^{spa}\left(\dot{v}^{t}_{i}\right)\right)\|_{2}
    ~\label{eq:spa_cons}
\end{equation}
\end{small}
where $\left\|\cdot\right\|_{2}$ denotes the $l2$ norm of the vector. $\emph{X}^{t}\left(\dot{v}^{t}_{i}\right)$ and $\emph{X}^{t}\left(\Tilde{S}^{spa}\left(\dot{v}^{t}_{i}\right)\right)$ are node feature matrices of the master node and the spatial candidate slave node set respectively. $\theta_{spa}$ is a specific trainable projection matrix when generating the spatial hyperedge $e_{spa}\left(\dot{v}_{i}^{t}\right)\in\mathcal{E}^{t}$. $\emph{p}^{spa}_{\dot{v}^{t}_{i}}\in\mathcal{R}^{\left(N-1\right)}$ denotes the trainable reconstruction coefficient vector for a specific spatial hyperedge, with each element $\emph{p}^{spa}_{\dot{v}^{t}_{i}}\left(v\right)$ representing the learned reconstruction coefficient of the node $v\in\Tilde{S}^{spa}\left(\dot{v}^{t}_{i}\right)$. According to $\emph{p}^{spa}_{\dot{v}^{t}_{i}}$, the nodes in the spatial candidate slave node set $\Tilde{S}^{spa}\left(\dot{v}^{t}_{i}\right)$ with reconstruction coefficient larger than zero are selected to generate a spatial hyperedge of the master node $\dot{v}^{t}_{i}$, which is denoted as $S^{spa}\left(\dot{v}^{t}_{i}\right) = \left\{v|v\in\Tilde{S}^{spa}\left(\dot{v}^{t}_{i}\right),\emph{p}^{spa}_{\dot{v}^{t}_{i}}\left(v\right)>0\right\}$ (connected through green solid lines in Fig.~\ref{fig:all}). By contrast, unselected nodes with a reconstruction coefficient value below 0 are connected by dotted lines. The spatial reconstruction error $c_{spa}$ measures the deviation of the reconstruction of the master node.

Similarly, as shown in the Fig.~\ref{fig:all}(a), temporal salve nodes are encircled from $\emph{V}^{t-1}$ to form the temporal hyperedge $e_{tem}\left(\dot{v}_{i}^{t}\right)$ with the master node based on the trainable reconstruction coefficient vector $\emph{p}^{tem}_{\dot{v}^{t}_{i}}$. 
Overall, the loss of hyperedges generation in one timestamp is defined as:
\begin{small}
\begin{equation}
\begin{split}
        \mathcal{L}_{recon}\!\! = \!\!\!\!\sum_{i=\left[1,...,N\right]}&\!\!\!\!\!\!\lambda\left(c_{spa}\!\left(\dot{v}^{t}_{i}\right)\!+\!c_{tem}\!\left(\dot{v}^{t}_{i}\right)\right)
        \!+\!\left(  \|\emph{p}^{spa}_{\dot{v}^{t}_{i}}\|_{1}+\|\emph{p}^{tem}_{\dot{v}^{t}_{i}}\|_{1}\right)\\&+\gamma\left(
    \|\emph{p}^{spa}_{\dot{v}^{t}_{i}}\|_{2}+\|\emph{p}^{tem}_{\dot{v}^{t}_{i}}\|_{2}\right)
    ~\label{eq:tem_cons}
\end{split}
\end{equation}
\end{small}
where $\left\|\cdot\right\|_{1}$ denotes the $l1$ norm of the vector and $c_{tem}$ denotes the reconstruction error of temporal hyperedges. $\lambda$ is the weight hyperparameter of the reconstruction error. $\gamma$ is the regularizing factor to balance $l1$ norm and $l2$ norm of the two types of reconstruction coefficient vectors. 
\subsection{Spatial-Temporal Hypergraph Learning}
\subsubsection{Hyperedge Embedding Updating}
Let $\emph{H}$ represents incidence relations between hyperedges and nodes, with entries are $ \emph{H}\left(v,e\left(\dot{v}^{t}_{i}\right)\right) =\left\{ \begin{small}
    \begin{matrix}
      1\text{,} & v = \dot{v}^{t}_{i} \\
      \emph{p}_{\dot{v}^{t}_{i}}\!\left(\!v\!\right)\!\text{,} & v \in S^{spa}\!\left(\!\dot{v}^{t}_{i}\!\right) \\
      0\text{,} & \text{otherwise}
    \end{matrix}\end{small}\right.$.
    The embedding of hyperedges is aggregated by node features as follows: 
\begin{small}    
\begin{equation}
    \emph{E}\left(e\left(\dot{v}^{t}_{i}\right)\right) = \frac{\sum_{v\in\mathcal{V}^{t}}\emph{H}\left(v,e\left(\dot{v}^{t}_{i}\right)\right)\times\emph{X}\left(v\right)}{\sum_{v\in\mathcal{V}^{t}}\emph{H}\left(v,e\left(\dot{v}^{t}_{i}\right)\right)}
    \label{eq:edge_emb}
\end{equation}
\end{small}
\subsubsection{Multi-head Node Embedding Updating}
We use the spatial hyperedge and the temporal hyperedge associated with the master node $\dot{v}^{t}_{i}$ to update the node embedding. The hyperedges associated with node $\dot{v}^{t}_{i}$ is denoted as $\left\{e_{spa}\left(\dot{v}^{t}_{i}\right),e_{tem}\left(\dot{v}^{t}_{i}\right)\right\}$. We calculate the multi-head attention between a master node and two types of hyperedge, and then use the normalized attention as the weight of the hyperedge. The process of weight calculation for two types of hyperedge is similar, and we thus take calculating spatial hyperedge attention as an example:
\begin{small}
\begin{equation}
\begin{split} 
    &\emph{Q}^{h}\left(\dot{v}^{t}_{i}\right)= \emph{X}^{t}\left(\dot{v}^{t}_{i}\right)\cdot Q\text{-}Lin^{h} \\
    &\emph{K}_{spa}^{h}\left(\dot{v}^{t}_{i}\right) = \emph{E}\left(e_{spa}\left(\dot{v}^{t}_{i}\right)\right)\cdot K\text{-}Lin^{h}_{spa}\\
    &att^{h}\left(\dot{v}^{t}_{i},e_{spa}\left(\dot{v}^{t}_{i}\right)\right) =  \frac{\emph{Q}^{h}\left(\dot{v}^{t}_{i}\right)\cdot\Theta_{spa}^{att}\cdot\emph{K}_{spa}^{h}\left(\dot{v}^{t}_{i}\right)^{T}}{\sqrt{d}}
\end{split}
\end{equation}
\end{small}
First, for the $h$-th attention head $att^{h}\left(\dot{v}^{t}_{i},e_{spa}\left(\dot{v}^{t}_{i}\right)\right)$, we project the master node into the $h$-th query vector $\emph{Q}^{h}\left(\dot{v}^{t}_{i}\right)$ with a linear transformation matrix $Q\text{-}Lin^{h}\in\mathcal{R}^{d\times\frac{d}{K}}$, where $K$ is the number of attention heads. We also project the spatial hyperedge $e_{spa}\left(\dot{v}^{t}_{i}\right)$ into the $h$-th key vector $\emph{K}_{spa}^{h}\left(\dot{v}^{t}_{i}\right)$ on the same dimension. Next, we apply a trainable weight matrix $\Theta_{spa}^{att}\in\mathcal{R}^{\frac{d}{K}\times\frac{d}{K}}$ to obtain $h$-th spatial attention, and $\sqrt{d}$ acts a scaling factor. The $h$-th temporal attention $att^{h}\left(\dot{v}^{t}_{i},e_{tem}\left(\dot{v}^{t}_{i}\right)\right)$ is calculated in a similar manner. Finally, the weight of the spatial hyperedge $w^{h}_{spa}\left(\dot{v}^{t}_{i}\right)$ and the temporal hyperedge $w^{h}_{tem}\left(\dot{v}^{t}_{i}\right)$ are calculated by $softmax$ normalization. The attentive aggregation of different heads among hyperedges for updating node embedding of $\dot{v}_{i}^{t}$ (reffered to as wavy lines in Fig ~\ref{fig:all}(b)) is denoted as :
\begin{small}    
\begin{equation}
\begin{split}    
    \emph{Z}_{\emph{V}^{t}}\left(\dot{v}^{t}_{i}\right) = MLP(\mathop{\concat}\limits_{h\in\left[1,K\right]}&(w^{h}_{spa}\left(\dot{v}^{t}_{i}\right)\times \emph{K}_{spa}^{h}\left(\dot{v}^{t}_{i}\right)\\&+w^{h}_{tem}\left(\dot{v}^{t}_{i}\right)\times \emph{K}_{tem}^{h}\left(\dot{v}^{t}_{i}\right)))
\end{split}
~\label{eq:att}
\end{equation}
\end{small}
where $\concat$ represents concatenation. We first aggregate two types of hyperedges associated with the master node, and then concatenate all $K$ heads. After that, the node embedding of $\dot{v}^{t}_{i}$ is updated by a shallow multi-layer perceptron (MLP). We average the node embedding of all nodes $\emph{Z}_{\emph{V}^{t}}$ at timestamp $t$ to read-out the graph representation of $\mathcal{G}^{t}$, which is denoted as $\emph{Z}_{\mathcal{G}^{t}}\in\mathcal{R}^{d}$.


To realise \algname in an end-to-end fashion, the loss function used in the training process is denoted as:
\begin{small}
\begin{equation}
    \mathcal{L} = \alpha\mathcal{L}_{recon}+\left(1-\alpha\right)CE\left(MLP\left(\emph{Z}_{\mathcal{G}^{t}}\right),\emph{y}^{t}\right)
\end{equation}
\end{small}
 The $MLP$ is used to map from graph representation to label space, and the cross-entropy function (CE) measures the difference between predicted labels and true labels of graphs. $\alpha$ is a weight hyperparameter to trade off the effects of reconstruction loss and the loss of graph classification.

%% file: 4_experiment.tex
\section{experiments}
\label{sec:intro}
\subsection{Experimental Setting}
The evaluation of \algname is performed based on the ISRUC Subgroup-3 dataset~\cite{khalighi2016isruc}, which encompasses PSG recordings collected from 10 healthy subjects. The time-series data is extracted by decomposing each PSG recording into 6 EEG channels, 2 EOG channels, 1 EMG channel and 1 ECG channel. To make a fair comparison with other state-of-the-art baselines, all 10 channels are employed as the input into our model. While the raw signal contains a feature dimension of 6000, we use a Covolutational Neural Network ~\cite{GraphSleepNet} for feature extraction, reducing the dimension down to 256. 

The proposed model is compared with five baseline methods that fall into traditional and graph-based categories. Support Vector Machine (SVM)~\cite{HASSAN2016107} is a traditional machine learning method that learns the classification boundary by optimizing the hinge loss. Random Forest (RF)~\cite{RF} introduces the bagging strategy to perform classification, with the help of multiple decision trees. DeepSleepNet~\cite{supratak2017deepsleepnet} is a deep learning based method that encodes spatial signal information using convolution neural network and temporal signal information with the Bi-LSTM network.
Multi-View Spatial-Temporal Graph Convolution Network (MSTGCN)~\cite{MSTGCN} is a graph learning based method that employs the attention mechanism and spatial-temporal network to capture the signal information.
SleepHGNN~\cite{SleepHGNN} is a graph learning based method that constructs a heterogeneous graph based on the mutual information between channels, Heterogeneous Graph Transformer is further applied to perform graph-level classification. We follow the data pre-processing steps and the default model structures reported in the original papers for all the baselines introduced above. For SVM, RF, DeepSleepNet and \algname, we provide spatial-temporal data from two consecutive timestamps $\left\{\emph{X}^{t-1},\emph{X}^{t}\right\}$, while the data for MSTGCN is provided within a temporal window with size of 5, i.e., $\left\{\emph{X}^{t-2},\emph{X}^{t-1}, \emph{X}^{t},\emph{X}^{t+1},\emph{X}^{t+2}\right\}$, in accordance with its reported experimental setting. Feature extraction network is applied for dimensional reduction for both MSTGCN and \algname. Regarding the hyperparameters in the reconstruction loss function, $\lambda$ is set to 0.01, $\gamma$ to 0.2 and $\alpha$ to 0.1, respectively.

\subsection{Experimental Results}
The experimental results are summarised in Table~\ref{BR}, with weighted F1-score and Accuracy applied as the evaluation metrics. On average, \algname outperforms the traditional models with 15.4\% average performance gain in F1 score and 12.1\% in Accuracy, respectively. It indicates that the traditional models are incapable of simultaneously capturing spatial information and temporal information due to the simpleness of model structures, while \algname updates node embeddings based on both spatial and temporal hyperedges, resulting in better modelling capability for spatial-temporal data. Compared with graph-based models, \algname achieves an average performance gain of 2.9\% in F1 score and 3.7\% in Accuracy. MSTGCN and SleepHGNN can only perform pairwise message passing between adjacent nodes, whereas \algname uncovers the implicit information through the constructed hyperedges.

\begin{table}[]
\centering
\resizebox{6cm}{!}{%
\begin{tabular}{|c|c|cc|}
\hline
\multicolumn{2}{|c|}{Models}                                                                  & F1    & Accuracy \\ \hline
\multirow{3}{*}{\begin{tabular}[c]{@{}c@{}}Traditional\\ Model\end{tabular}} & SVM          & 0.653 & 0.720    \\
                                                                             & RF           & 0.612 & 0.699    \\
                                                                             & DeepSleepNet & 0.766 & 0.767    \\ \hline
\multirow{3}{*}{\begin{tabular}[c]{@{}c@{}}Graph Based\\ Model\end{tabular}} & MSTGCN       & 0.824 & 0.829    \\
                                                                             & SleepHGNN    & 0.779 & 0.795    \\
                                                                             & \algname   & \textbf{0.831} & \textbf{0.849}   \\ \hline
\end{tabular}%
}
\captionsetup{justification=centering}
\caption{Evaluation Result}
\label{BR}
\end{table}

\begin{table}[]
\centering
\vspace{-0.2cm}
\resizebox{5cm}{!}{%
\begin{tabular}{|c|cc|}
\hline
Model Setting & F1    & Accuracy \\ \hline
Default       & 0.849 & 0.831    \\ \hline
w/o Hyperedge & 0.782 & 0.774    \\ \hline
w/o Attention & 0.818 & 0.806    \\ \hline
\end{tabular}%
}
\captionsetup{justification=centering}
\caption{Ablation Study Result}
\label{tab:my-table}
\end{table}

The performance gain of \algname is mainly contributed by the hyperedge reconstruction and the multi-head attentive node embedding updating. Two model variants are delineated in Table~\ref{tab:my-table} for further analysis. We first study the effect without using the hyperedges, in which the vanilla graph convolution layer is utilised to update nodes with spatial information. Note the adjacency matrix is calculated using Pearson coefficients between node embeddings to quantify the connectivity. Nodes are updated with temporal information based on the weighted sum of node embeddings from previous and current timestamps. Removing hyperedges results in a 6.7\% performance drop in F1 score and a 5.7\% drop in Accuracy due to the elimination of learnable reconstruction parameters. The second variant of the model relates to removing the multi-head attentive layer when updating node embeddings. The variant results in a 3.1\% drop in F1 score and a 2.5\% drop in Accuracy, inferring the necessity of modelling spatial and temporal hyperedges separately with attention weights.

%% file: 5_conclusion.tex
\section{conclusion}
\label{sec:con}
We propose a novel hypergraph model for sleep stage classification, which contains dynamic hyperedges generation and spatial-temporal hypergraph learning. The node embedding is updated through type-specific attention, aiming to encode the heterogeneity and interactivity into the embedding space. Experimental results demonstrate that \algname outperforms the state-of-the-art models on the sleep stage classification.